\documentclass[sigconf]{acmart}

\usepackage{amsmath}
\usepackage{array}
\usepackage{booktabs}
\usepackage{multirow}

\setcopyright{none}
\acmConference[WWW '27]{The ACM Web Conference 2027}{May 10--14, 2027}{Dublin, Ireland}
\acmYear{2027}
\acmISBN{978-x-xxxx-xxxx-x/YYYY/MM}
\acmDOI{10.1145/nnnnnnn.nnnnnnn}

\newcommand{\method}{UMC}
\newcommand{\cplr}{CPLR}
\newcommand{\rpc}{RPC}
\newcommand{\R}{\mathbb{R}}
\newcommand{\clip}{\operatorname{clip}}

\begin{document}

\title{Uncertainty-Guided Multimodal Consensus for Multimodal Sentiment Analysis}

\author{Guowei Zhong}
\email{gwzhong@zju.edu.cn}
\affiliation{%
  \institution{Zhejiang University}
  \city{Hangzhou}
  \postcode{310027}
  \country{China}
}

\author{Junjie Li}
\email{lijunjie\_isee2019@zju.edu.cn}
\affiliation{%
  \institution{Zhejiang University}
  \city{Hangzhou}
  \postcode{310027}
  \country{China}
}
\author{Huaiyu Zhu}
\email{zhuhuaiyu@zju.edu.cn}
\affiliation{%
  \institution{Zhejiang University}
  \city{Hangzhou}
  \postcode{310027}
  \country{China}
}

\author{Ruohong Huan}
\email{huanrh@zjut.edu.cn}
\affiliation{%
  \institution{Zhejiang University of Technology}
  \city{Hangzhou}
  \postcode{310023}
  \country{China}
}

\author{Yun Pan}
\authornote{Corresponding author.}
\email{panyun@zju.edu.cn}
\affiliation{%
  \institution{Zhejiang University}
  \city{Hangzhou}
  \postcode{310027}
  \country{China}
}

\renewcommand{\shortauthors}{Zhong et al.}

\begin{abstract}
Multimodal sentiment analysis (MSA) has made considerable progress in recent years. Existing MSA methods mainly learn shared multimodal representations or train unimodal predictors with a common multimodal label. However, assigning the shared label to every modality can obscure modality-specific sentiment, while direct fusion may overemphasize unreliable predictions when modalities disagree. To {address} these issues, we propose Uncertainty-Guided Multimodal Consensus (UMC), which first recovers continuous unimodal pseudo-labels and then forms a reliability-aware consensus across modalities. Our method consists of two key components: i) Continuous Prototype-Guided Label Recovery (CPLR), which combines heteroscedastic predictions with ordered sentiment prototypes and uses confidence and label conflict to refine continuous unimodal pseudo-labels, and ii) Reliability and Precision Consensus (RPC), which estimates sample-dependent modality reliability from predictive statistics and cross-modal disagreement and combines it with predictive precision through a closed-form consensus solver. Extensive experiments conducted on four publicly available MSA datasets demonstrate the effectiveness and superiority of our proposed model.
\end{abstract}


\maketitle

\section{Introduction}

\begin{figure*}[t]
  \centering
  \includegraphics[width=0.82\textwidth]{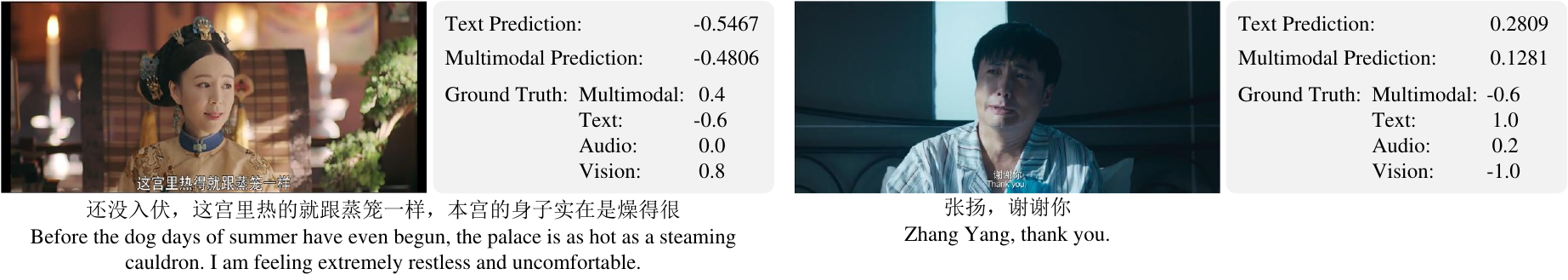}
  \caption{Cross-modal semantic conflict in two CH-SIMS examples. Text-only and multimodal predictions from EMT-DLFR~\protect\cite{sun2023efficient} retain the polarity of the textual annotation despite conflicting nonverbal annotations. Labels range from $-1$ (negative) to $1$ (positive).}
  \Description{Two CH-SIMS video examples compare independent modality annotations with text-only and multimodal predictions from EMT-DLFR.}
  \label{fig:motivation}
\end{figure*}

The rapid growth of opinion videos on social platforms has increased interest in multimodal sentiment analysis (MSA), which predicts sentiment from language, vocal delivery, and facial behavior~\cite{zadeh2017tensor,liu2018efficient,tsai2019multimodal}. Combining these signals supports Web video understanding, but remains difficult when the modalities express different sentiments~\cite{yu2020ch,liu2022make,luo2025triagedmsa}. Two problems arise in this setting. First, a single multimodal annotation does not directly describe the continuous sentiment conveyed by each modality. Second, direct fusion may allow a strong but misleading modality prediction to dominate the output. Existing fusion architectures mainly exchange or aggregate modality information and optimize fused predictions using multimodal annotations, without explicitly recovering modality-specific training targets~\cite{zadeh2017tensor,liu2018efficient,tsai2019multimodal,hazarika2020misa}. Such supervision leaves each modality's sentiment unspecified when it differs from the multimodal annotation. As illustrated in Figure~\ref{fig:motivation}, the textual annotation in two CH-SIMS~\cite{yu2020ch} utterances has the opposite polarity to the multimodal annotation and at least one nonverbal annotation. The corresponding text-only and multimodal EMT-DLFR~\cite{sun2023efficient} predictions retain the textual polarity and miss the multimodal target. These examples show why modality-specific sentiment and uncertainty should be estimated before fusion.

Although recent studies improve cross-modal interaction, mo\-dality-specific learning, and adaptive weighting~\cite{yu2021learning,wang2025dlf,yang2025mse,gao2024enhanced}, they usually {focus on} only one part of this process. Pseudo-label methods construct modality-level supervision from a {multimodal annotation}, but do not jointly use input-depen\-dent uncertainty to determine how a continuous target should depart from that annotation~\cite{yu2021learning,zhu2025multimodal}. Conversely, learned routing methods adaptively route expert features or reweight modality representations without explicitly refining modality-specific training targets~\cite{gao2024enhanced,ma2026mruf}. The two problems are closely connected because inaccurate modality targets produce unreliable experts, while direct fusion can amplify their errors when predictions disagree. Therefore, how to recover continuous mo\-dality-specific supervision and then form a reliable consensus under cross-modal disagreement remains an open problem.

To {address} this problem, we propose \method{}, a two-stage uncer\-tainty-guided framework that separates modality-specific label recovery from consensus formation. We first develop Continuous Prototype-Guided Label Recovery (\cplr{}), which pretrains three unimodal regression experts to predict a sentiment mean and an input-depen\-dent variance. Each expert also retrieves a second sentiment estimate from ordered prototypes, after which confidence and conflict with the multimodal annotation guide the continuous pseudo-label update. We then introduce Reliability and Precision Consensus (\rpc{}), where a shared router estimates modality reliability from representations, predictions, uncertainty, pairwise disagreement, and modality identity. A closed-form weighted least-squares solver combines the learned reliability scores with predictive precision to obtain the final consensus. Extensive experiments on four publicly available MSA datasets demonstrate the effectiveness of both components and the overall model.

The main contributions of this work are threefold:

\begin{itemize}
  \item We introduce \cplr{}, which combines input-dependent uncertainty with ordered continuous prototypes to recover modality-specific pseudo labels from shared multimodal supervision.
  \item We design \rpc{}, which estimates reliability from modality predictions and their disagreement, then combines reliability with predictive precision through an explicit closed-form objective.
  \item We evaluate \method{} on four MSA benchmarks and demonstrate its effectiveness and superiority over the compared methods.
\end{itemize}

\section{Related Work}

MSA aims to capture sentiment information from multimodal data, such as text, audio, and visual behavior, and use this information to predict sentiment. {We organize related work into four categories according to the main mechanisms of existing methods.}

{
\subsection{Multimodal Representation Learning and Fusion}

Multimodal representation and fusion methods integrate heterogeneous modality features through distinct interaction mechanisms. MulT applies directed cross-modal attention to unaligned sequences~\cite{tsai2019multimodal}. DMD transfers knowledge between decoupled modality subspaces~\cite{li2023decoupled}, while DLF disentangles language-focused representations~\cite{wang2025dlf}. These methods improve feature interaction or representation separation for sentiment prediction.

These methods do not explicitly recover modality-specific training targets. UMC instead uses \cplr{} to recover continuous unimodal targets from multimodal annotations, and \rpc{} to combine expert predictions according to predictive uncertainty and cross-modal disagreement.
\par}

{
\subsection{Pseudo-Label Generation}

Pseudo-label generation methods infer modality-specific sentiment targets when unimodal annotations are unavailable. To obtain modality-specific supervision from multimodal labels, Self-MM generates unimodal supervision signals and jointly optimizes unimodal and multimodal tasks~\cite{yu2021learning}. SUGRM generates unimodal labels using cross-modally recalibrated representations~\cite{hwang2023self}. MUG uses meta-learning and denoising tasks to learn unimodal label corrections and evaluate their quality~\cite{mai2024meta}, while ULMD combines modality decomposition with label generation based on feature similarities to sentiment centers~\cite{zhu2025multimodal}.

Although these methods recover unimodal supervision, their feature-based or meta-learned corrections do not explicitly use predictive variance to control target refinement. In contrast, our \cplr{} module derives confidence from predictive uncertainty and prototype dispersion, and uses this confidence together with label conflict to guide continuous pseudo-label refinement.
\par}

{
\subsection{Semantic Inconsistency Modeling}

Semantic inconsistency methods explicitly model differences in the sentiment expressed by different modalities. TriagedMSA infers polarity-relation labels when unimodal annotations are unavailable and uses them to triage samples into different attention networks~\cite{luo2025triagedmsa}. RMA combines knowledge injection with distribution-based ambiguity measures to regulate language-guided cross-modal interactions~\cite{zhao2025resolving}, while CACR uses conflict-dependent weights for adaptive cross-reconstruction~\cite{wang2026conflict}.

Although these methods model semantic inconsistency through sample triage, cross-modal interaction, or reconstruction, they do not explicitly recover continuous unimodal targets when only multimodal annotations are available. In contrast, our UMC framework uses \cplr{} to refine continuous unimodal targets according to label conflict and predictive confidence, providing supervision that can better reflect each modality's sentiment.
\par}

{
\subsection{Expert Coordination and Uncertainty-Aware Fusion}

Expert-based methods adapt how mo\-dality-specific information contributes to the multimodal prediction. EMOE constructs mo\-dality-specific dynamic experts~\cite{fang2025emoe}. Uncertainty-aware methods further estimate the reliability of modality information. EUAR places uncer\-tainty-aware experts inside a mixture-of-experts router~\cite{gao2024enhanced}, while the closely related concurrent MRUF combines learned routing with inverse-variance calibration and leave-one-out importance supervision~\cite{ma2026mruf}. TMSON models unimodal feature distributions, combines them through Bayesian fusion, and introduces ordinal sentiment constraints~\cite{xie2024trustworthy}.

These methods improve expert coordination or uncertainty-aware fusion, but do not explicitly recover unimodal sentiment targets from multimodal annotations. UMC uses uncertainty-guided target recovery in \cplr{} before combining disagreement-aware reliability and predictive precision in \rpc{}.
\par}

\begin{figure*}[t]
  \centering
  \includegraphics[height=2.50in]{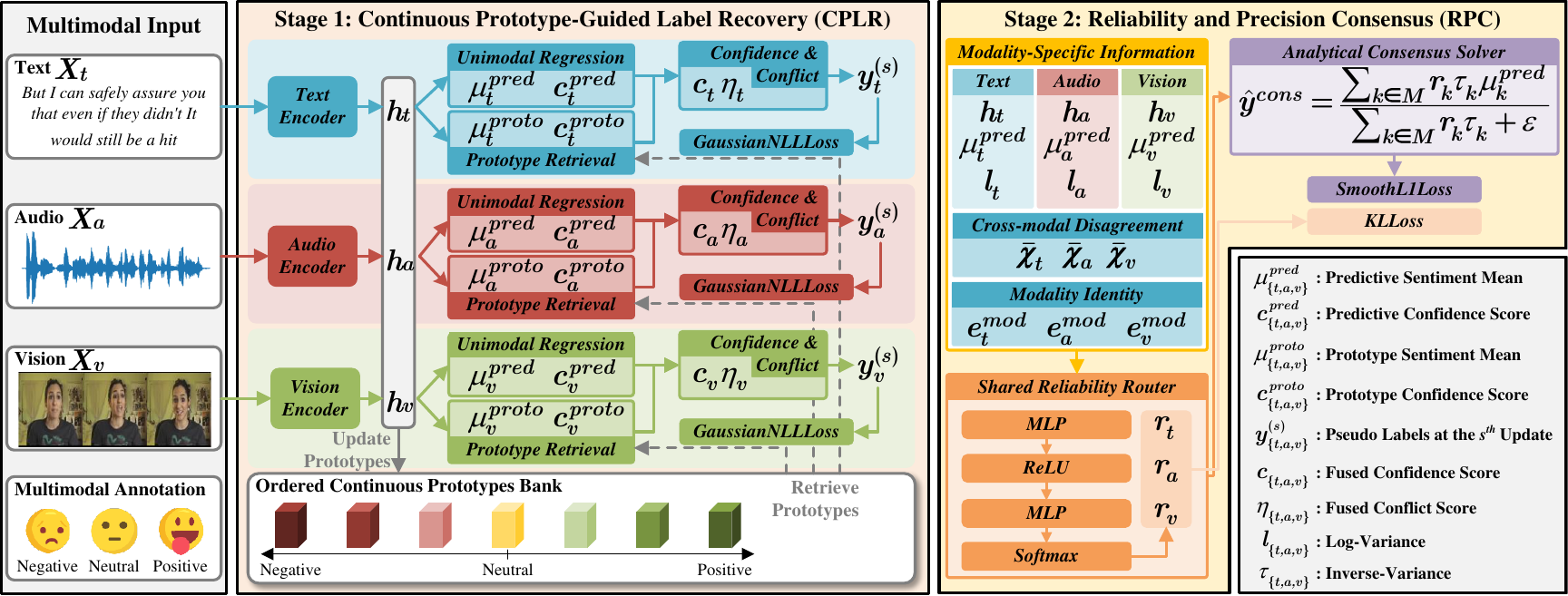}
  \caption{Overview of UMC with CPLR for continuous unimodal pseudo-label recovery and RPC for reliability and precision consensus.}
  \Description{Overview of UMC, including multimodal inputs, CPLR, and RPC.}
  \label{fig:legacy-framework}
\end{figure*}

\section{Method}
\label{sec:method}

This section presents the formulation and two stages of \method{}, as illustrated in Figure~\ref{fig:legacy-framework}. We first formulate the MSA task. Then, \cplr{} recovers continuous unimodal pseudo labels and pretrains three uncertainty-aware regression experts. Finally, \rpc{} transfers the experts and combines learned reliability with predictive precision to form the multimodal consensus.

\subsection{Problem Formulation}

Given utterance $i$, we denote the input sequence of modality $k\in\mathcal{M}=\{t,a,v\}$ as $X_{k,i}$ and its multimodal sentiment label as $y_i^m$. Here, $t$, $a$, and $v$ denote text, audio, and vision, and $m$ denotes the multimodal annotation. The task is to predict $y_i^m$ while retaining modality-specific sentiment and uncertainty estimates that expose disagreement among the modalities.

\subsection{Continuous Prototype-Guided Label Recovery (\cplr{})}

\cplr{} {aims to address} the supervision mismatch caused by assigning one multimodal annotation to every modality. It combines a heteroscedastic prediction with a continuous estimate from ordered sentiment prototypes. Confidence and label conflict then govern each pseudo-label update.

\subsubsection{Unimodal Predictive Distribution}

Each modality expert maps its input sequence to a temporally averaged and $\ell_2$-normalized representation $h_{k,i}$:
\begin{equation}
  h_{k,i}=f_k(X_{k,i})\in\R^{d_h},
\label{eq:unimodal-encoding}
\end{equation}
where $f_k$ is the encoder for modality $k$ and $d_h$ is the representation dimension.

The regression head refines $h_{k,i}$ into a residual feature $\tilde h_{k,i}$:
\begin{equation}
  \tilde h_{k,i}=h_{k,i}+W_{k,2}\operatorname{Dropout}_{\delta_k}\!\left(
  \operatorname{ReLU}(W_{k,1}h_{k,i}+b_{k,1})\right)+b_{k,2},
\label{eq:residual-projection}
\end{equation}
where $W_{k,1}$ and $W_{k,2}$ are projection matrices, $b_{k,1}$ and $b_{k,2}$ are biases, and $\operatorname{Dropout}_{\delta_k}$ applies modality-specific dropout with rate $\delta_k$ after $\operatorname{ReLU}$ activation.

The refined feature is used to predict the sentiment mean $\mu^{\mathrm{pred}}_{k,i}$ and log-variance $\ell_{k,i}$:
\begin{equation}
  (\mu^{\mathrm{pred}}_{k,i},\ell_{k,i})
  =g_k(h_{k,i})=\left((w_k^\mu)^\top\tilde h_{k,i}+b_k^\mu,
  (w_k^\ell)^\top\tilde h_{k,i}+b_k^\ell\right),
\label{eq:predictive-head}
\end{equation}
where $g_k$ is the complete regression head, $w_k^\mu$ and $w_k^\ell$ are output vectors, $b_k^\mu$ and $b_k^\ell$ are output biases, and $\top$ denotes transpose.

The predicted log-variance is converted into predictive variance $\sigma^2_{k,i}$:
\begin{equation}
  \sigma^2_{k,i}=\exp(\ell_{k,i}),
\label{eq:predictive-variance}
\end{equation}
where $\exp(\cdot)$ is the exponential function.

The predictive mean provides the modality-specific sentiment estimate, while
the variance expresses input-dependent uncertainty. CPLR uses this uncertainty
to recover pseudo labels, and RPC later converts the same quantity into fusion
precision.

\subsubsection{Ordered Continuous Prototype Bank}

To complement the direct prediction with sentiment-scale structure, \cplr{} maintains an ordered continuous prototype bank $\mathcal{B}$ containing $K$ anchors on the sentiment axis:
\begin{equation}
  \mathcal{B}=\{\kappa_1,\ldots,\kappa_K\},\qquad
  \kappa_j\in[y_{\min},y_{\max}],
\label{eq:anchor-bank}
\end{equation}
where $\kappa_j$ is anchor $j\in\{1,\ldots,K\}$ within the sentiment range $[y_{\min},y_{\max}]$.

The anchor spacing $T_y$ defines the bandwidth for soft assignment:
\begin{equation}
  T_y=\frac{y_{\max}-y_{\min}}{K-1}.
\label{eq:anchor-spacing}
\end{equation}

Given a continuous pseudo target $y$, we use a Gaussian kernel to compute its soft contribution $\pi_j(y)$ to anchor $j$:
\begin{equation}
  \pi_j(y)=
  \frac{\exp\!\left[-(\kappa_j-y)^2/(2T_y^2)\right]}
       {\sum_{q=1}^{K}\exp\!\left[-(\kappa_q-y)^2/(2T_y^2)\right]+\epsilon},
\label{eq:soft-bin}
\end{equation}
where $\epsilon=10^{-8}$ stabilizes the denominator. This assignment preserves the continuous relation between the target and the ordered anchors.

At update $s$, the bank aggregates the mini-batch features assigned to each modality-anchor pair into the weighted mean $\bar h_{k,j}$:
\begin{equation}
  \bar h_{k,j}=
  \frac{\sum_{i\in\mathcal{Q}}\pi_j(y_{k,i}^{(s-1)})h_{k,i}}
       {\sum_{i\in\mathcal{Q}}\pi_j(y_{k,i}^{(s-1)})+\epsilon},
\label{eq:prototype-batch-mean}
\end{equation}
where $\mathcal{Q}$ is the mini-batch and $y_{k,i}^{(s-1)}$ is the stored pseudo label from the preceding update.

Each zero-initialized prototype $C_{k,j}\in\R^{d_h}$ incorporates the corresponding mini-batch estimate through an exponential moving average:
\begin{equation}
  C_{k,j}\leftarrow \zeta C_{k,j}+(1-\zeta)\bar h_{k,j},
\label{eq:prototype-ema}
\end{equation}
where $\zeta\in[0,1)$ is the EMA momentum and $\leftarrow$ denotes an in-place update. The pseudo labels are initialized as $y_{k,i}^{(0)}=y_i^m$. Detached expert outputs update both the prototype and pseudo-label banks, while the updated labels supervise the predictions from the same forward pass.

After updating the bank, we convert the squared distances between the current feature and all prototypes into normalized affinities $\phi_{k,i,j}$:
\begin{equation}
  \phi_{k,i,j}
    =\frac{\exp(-\lVert h_{k,i}-C_{k,j}\rVert_2^2)}
       {\sum_{q=1}^{K}\exp(-\lVert h_{k,i}-C_{k,q}\rVert_2^2)},
\label{eq:prototype-affinity}
\end{equation}
where $\lVert\cdot\rVert_2^2$ is squared Euclidean distance.

The normalized affinities interpolate the discrete anchors into a continuous prototype estimate $\mu^{\mathrm{proto}}_{k,i}$:
\begin{equation}
  \mu^{\mathrm{proto}}_{k,i}
    =\sum_{j=1}^{K}\phi_{k,i,j}\kappa_j.
\label{eq:prototype-mean}
\end{equation}

To characterize the concentration of the retrieved anchors, we compute the normalized dispersion-based prototype-uncertainty proxy $u^{\mathrm{proto}}_{k,i}$:
\begin{equation}
  u^{\mathrm{proto}}_{k,i}
    =\frac{\sum_{j=1}^{K}\phi_{k,i,j}
      (\kappa_j-\mu^{\mathrm{proto}}_{k,i})^2}
      {((y_{\max}-y_{\min})/2)^2+\epsilon}.
\label{eq:prototype-dispersion}
\end{equation}

The prototype branch supplies a second estimate tied to the order of the
sentiment scale. Its dispersion increases when affinity is distributed across
distant anchors, allowing CPLR to distinguish a concentrated match from an
ambiguous retrieval.

\subsubsection{Confidence and Conflict Guided Update}

The predictive and prototype branches can contribute differently to pseudo-label recovery. We therefore estimate the confidence of each branch and measure the conflict between their fused estimate and the multimodal annotation. First, predictive variance is converted into the confidence score $c^{\mathrm{pred}}_{k,i}$:
\begin{equation}
  c^{\mathrm{pred}}_{k,i}=\exp(-\sigma^2_{k,i}).
\label{eq:prediction-confidence}
\end{equation}

We convert prototype dispersion into the prototype confidence score $c^{\mathrm{proto}}_{k,i}$ in the same manner:
\begin{equation}
  c^{\mathrm{proto}}_{k,i}=\exp(-u^{\mathrm{proto}}_{k,i}).
\label{eq:prototype-confidence}
\end{equation}

The two sentiment estimates are fused according to their respective confidence scores to produce $\mu^{\mathrm{cplr}}_{k,i}$:
\begin{equation}
  \mu^{\mathrm{cplr}}_{k,i}
    =\frac{c^{\mathrm{pred}}_{k,i}\mu^{\mathrm{pred}}_{k,i}
      +c^{\mathrm{proto}}_{k,i}\mu^{\mathrm{proto}}_{k,i}}
      {c^{\mathrm{pred}}_{k,i}+c^{\mathrm{proto}}_{k,i}+\epsilon}.
\label{eq:cplr-fusion}
\end{equation}

The average of the two branch confidences defines the overall confidence $c_{k,i}$ used for pseudo-label recovery:
\begin{equation}
  c_{k,i}=\frac{c^{\mathrm{pred}}_{k,i}+c^{\mathrm{proto}}_{k,i}}{2}.
\label{eq:cplr-confidence}
\end{equation}

We next normalize the disagreement between the fused estimate and the multimodal annotation to obtain label conflict $\eta_{k,i}\in[0,1]$:
\begin{equation}
  \eta_{k,i}=\clip\!\left(
  \frac{|\mu^{\mathrm{cplr}}_{k,i}-y_i^m|}{y_{\max}-y_{\min}},0,1\right),
\label{eq:label-conflict}
\end{equation}
where $|\cdot|$ denotes absolute value and
\newline$\clip(x,l,u)=\min(\max(x,l),u)$ bounds $x$ to $[l,u]$.

Confidence and label conflict jointly determine the pseudo-label update strength $\lambda_{k,i}\in[0,1]$:
\begin{equation}
  \lambda_{k,i}=\clip(c_{k,i}\eta_{k,i},0,1).
\label{eq:update-strength}
\end{equation}

Using this strength, we move the multimodal annotation toward the fused \cplr{} estimate and obtain the bounded candidate pseudo label $\tilde y_{k,i}$:
\begin{equation}
  \tilde y_{k,i}=\clip\!\left(
  y_i^m+\lambda_{k,i}(\mu^{\mathrm{cplr}}_{k,i}-y_i^m),
  y_{\min},y_{\max}\right).
\label{eq:candidate-label}
\end{equation}

Finally, we smooth the candidate with the preceding bank value to obtain the new pseudo label $y_{k,i}^{(s)}$:
\begin{equation}
  y_{k,i}^{(s)}=\gamma_{k,i}y_{k,i}^{(s-1)}
  +(1-\gamma_{k,i})\tilde y_{k,i},
\label{eq:pseudo-ema}
\end{equation}
where $\gamma_{k,i}=1-c_{k,i}$ is the history coefficient.

Conflict determines whether the {multimodal annotation} should move toward the
modality-specific estimate, and confidence controls the update magnitude. The
history coefficient smooths these revisions across bank updates.

\subsubsection{Stage I Optimization}

The bank update does not propagate gradients. After this update, we optimize each expert against its new pseudo target using the scaled Gaussian negative log-likelihood $\mathcal{L}^{\mathrm{uni}}_k$~\cite{nix1994estimating}:
\begin{equation}
  \mathcal{L}^{\mathrm{uni}}_k
  =\frac{1}{N}\sum_{i=1}^{N}\left[
  \exp(-\ell_{k,i})(\mu^{\mathrm{pred}}_{k,i}-y_{k,i}^{(s)})^2
  +\ell_{k,i}\right],
\label{eq:unimodal-loss}
\end{equation}
where $N=|\mathcal{Q}|$ is the current mini-batch size.

The complete Stage I objective $\mathcal{L}_{\mathrm{stage1}}$ sums the losses of the three modality experts:
\begin{equation}
  \mathcal{L}_{\mathrm{stage1}}
  =\sum_{k\in\mathcal{M}}\mathcal{L}^{\mathrm{uni}}_k.
\label{eq:stage1-loss}
\end{equation}

\subsection{Reliability and Precision Consensus (\rpc{})}

The three \cplr{} experts need not be equally dependable for every sample. \rpc{} therefore estimates sample-dependent reliability with a shared router and combines it with variance-derived precision. Their product determines each expert's contribution to the consensus.

\subsubsection{Uncer\-tainty- and Dis\-agreement-Aware Reliability Router}

\leavevmode\newline Stage II discards the prototype branch and transfers $h_{k,i}$, $\mu^{\mathrm{pred}}_{k,i}$, and $\ell_{k,i}$ from the three regression experts. We first convert predictive variance into precision $\tau_{k,i}$:
\begin{equation}
  \tau_{k,i}=\frac{1}{\sigma^2_{k,i}+\epsilon}.
\label{eq:precision}
\end{equation}

We then measure the normalized prediction disagreement $\chi_{i,kq}$ between modalities $k$ and $q$:
\begin{equation}
  \chi_{i,kq}=\frac{|\mu^{\mathrm{pred}}_{k,i}-\mu^{\mathrm{pred}}_{q,i}|}
    {y_{\max}-y_{\min}},
\label{eq:pair-disagreement}
\end{equation}
where $q$ is the compared-modality index.

For each modality, the pairwise values are averaged to obtain the mean disagreement statistic $\bar\chi_{k,i}$:
\begin{equation}
  \bar\chi_{k,i}=\frac{1}{|\mathcal{M}|-1}
    \sum_{q\in\mathcal{M},q\ne k}\chi_{i,kq},
\label{eq:mean-disagreement}
\end{equation}
where $|\mathcal{M}|$ is the number of modalities.

The router input $z_{k,i}$ concatenates the transferred representation, predictive statistics, mean disagreement, and a modality embedding:
\begin{equation}
  z_{k,i}=\left[h_{k,i};\mu^{\mathrm{pred}}_{k,i};
    \ell_{k,i};\bar\chi_{k,i};e_k^{\mathrm{mod}}\right],
\label{eq:router-input}
\end{equation}
where $e_k^{\mathrm{mod}}$ is the learned modality embedding. The representation and predictive mean describe content and polarity, while log-variance and disagreement describe uncertainty and conflict. This input allows one shared scorer to compare heterogeneous experts while retaining modality identity and cross-modal disagreement.

A shared two-layer scorer $g_{\mathrm{r}}$ maps each router input to a scalar score $\xi_{k,i}$:
\begin{equation}
  \xi_{k,i}=g_{\mathrm{r}}(z_{k,i}),
\label{eq:router-score}
\end{equation}
where $g_{\mathrm{r}}$ is shared across the three modalities.

Finally, a softmax operation normalizes the scores across modalities to obtain sample-dependent reliability $r_{k,i}$:
\begin{equation}
  r_{k,i}=\frac{\exp(\xi_{k,i})}{\sum_{q\in\mathcal{M}}\exp(\xi_{q,i})}.
\label{eq:reliability}
\end{equation}

Reliability and precision serve different roles. The router learns comparative
expert quality from data, while precision reflects predictive uncertainty. RPC
retains both signals when forming the consensus.

\subsubsection{Analytical Consensus Solver}

To combine learned reliability with variance-derived precision, we define the consensus candidate $y_i^*$ as the minimizer of a weighted least-squares problem:
\begin{equation}
  y_i^*=\arg\min_{y\in\R}\sum_{k\in\mathcal{M}}
  r_{k,i}\tau_{k,i}(y-\mu^{\mathrm{pred}}_{k,i})^2,
\label{eq:consensus-objective}
\end{equation}
where $y\in\R$ is a candidate sentiment and $r_{k,i}\tau_{k,i}$ is the combined weight of expert $k$.

Solving this problem gives a weighted mean of the three predictions. The implemented consensus prediction $\hat y_i^{\mathrm{cons}}$ adds $\epsilon$ to the denominator for numerical stability and reduces to $y_i^*$ when this stabilizer is omitted:
\begin{equation}
  \hat y_i^{\mathrm{cons}}=
  \frac{\sum_{k\in\mathcal{M}}r_{k,i}\tau_{k,i}\mu^{\mathrm{pred}}_{k,i}}
       {\sum_{k\in\mathcal{M}}r_{k,i}\tau_{k,i}+\epsilon}
  \approx y_i^*.
\label{eq:consensus-prediction}
\end{equation}

\subsubsection{Stage II Training Objective}

The Stage II objective fits the consensus, guides the router toward accurate experts, and preserves Stage I supervision. First, we measure prediction error using the consensus loss $\mathcal{L}_{\mathrm{cons}}$:
\begin{equation}
  \mathcal{L}_{\mathrm{cons}}=
  \frac{1}{N}\sum_{i=1}^{N}
  \operatorname{SmoothL1}(\hat y_i^{\mathrm{cons}},y_i^m),
\label{eq:consensus-loss}
\end{equation}
where $\operatorname{SmoothL1}$ uses the default unit transition point~\cite{girshick2015fast}.

To supervise the router directly, we convert detached unimodal errors into the error-based reliability target $\tilde r_{k,i}$:
\begin{equation}
  \tilde r_{k,i}=
  \frac{\exp(-|\operatorname{sg}[\mu^{\mathrm{pred}}_{k,i}]-y_i^m|/T_r)}
       {\sum_{q\in\mathcal{M}}\exp(-|\operatorname{sg}[\mu^{\mathrm{pred}}_{q,i}]-y_i^m|/T_r)},
\label{eq:oracle-reliability}
\end{equation}
where $\operatorname{sg}[\cdot]$ denotes stop-gradient and $T_r>0$ is the target temperature.

We train the router with the batch-mean divergence $\mathcal{L}_{\mathrm{router}}$ between the target and predicted reliability distributions:
\begin{equation}
  \mathcal{L}_{\mathrm{router}}=
  \frac{1}{N}\sum_{i=1}^{N}\sum_{k\in\mathcal{M}}
  \tilde r_{k,i}\log\frac{\tilde r_{k,i}}{r_{k,i}+\epsilon},
\label{eq:router-loss}
\end{equation}
where $\log$ denotes the natural logarithm.

To preserve the modality-specific supervision learned in Stage I, each expert retains the Stage II heteroscedastic regularizer $\mathcal{L}^{\mathrm{uni,II}}_k$ based on the pseudo label saved at its best Stage I checkpoint:
\begin{equation}
  \mathcal{L}^{\mathrm{uni,II}}_k=
  \frac{1}{N}\sum_{i=1}^{N}\left[
  \exp(-\ell_{k,i})(\mu^{\mathrm{pred}}_{k,i}-\bar y_{k,i})^2+\ell_{k,i}
  \right],
\label{eq:stage2-unimodal-loss}
\end{equation}
where $\bar y_{k,i}$ is the fixed pseudo label from the best modality-specific Stage I checkpoint.

Finally, the consensus, router, and unimodal terms form the complete Stage II objective $\mathcal{L}$:
\begin{equation}
  \mathcal{L}=\mathcal{L}_{\mathrm{cons}}
  +\alpha\mathcal{L}_{\mathrm{router}}
  +\beta\sum_{k\in\mathcal{M}}\mathcal{L}^{\mathrm{uni,II}}_k,
\label{eq:total-loss}
\end{equation}
where $\alpha,\beta\ge0$ weight router supervision and unimodal regularization. The experts and router are optimized jointly.

\section{Experiments}
\label{sec:experiments}

\begin{table*}[t]
  \centering
  \caption{Comparative experiments with the best-performing MSA models conducted on the CMU-MOSI, CMU-MOSEI, CH-SIMS, and CH-SIMS v2 datasets. Except for MAE ($\downarrow$), higher values ($\uparrow$) indicate better performance. Bold and underlined values are best and second best. Starred results are reproduced from released code in the same environment as UMC. Unstarred results for CMU-MOSI and CMU-MOSEI come from DLF~\protect\cite{wang2025dlf}. Those for CH-SIMS and CH-SIMS v2 come from the \href{https://github.com/thuiar/MMSA/blob/master/results/result-stat.md}{MMSA repository}~\protect\cite{mao2022m}.}
  \label{tab:main-results}

  \resizebox{\textwidth}{!}{%
  \begin{tabular}{lrrrrrrrrrrrr}
    \toprule
    & \multicolumn{6}{c}{CMU-MOSI} & \multicolumn{6}{c}{CMU-MOSEI} \\
    \cmidrule(lr){2-7}\cmidrule(lr){8-13}
    Method & Acc-7$\uparrow$ & Acc-5$\uparrow$ & Acc-2$\uparrow$ & F1$\uparrow$ & MAE$\downarrow$ & Corr$\uparrow$
    & Acc-7$\uparrow$ & Acc-5$\uparrow$ & Acc-2$\uparrow$ & F1$\uparrow$ & MAE$\downarrow$ & Corr$\uparrow$ \\
    \midrule
    MulT~\cite{tsai2019multimodal} & 40.00 & 42.68 & 83.00 & 82.00 & 0.8710 & 0.6980 & 51.80 & 54.18 & 82.50 & 82.30 & 0.5800 & 0.7030 \\
    Self-MM$^*$~\cite{yu2021learning} & \underline{47.67} & \underline{54.37} & 83.23 & 83.25 & 0.7267 & 0.7813 & \underline{54.20} & \underline{55.87} & 85.17 & 85.16 & \underline{0.5356} & \underline{0.7690} \\
    {SUGRM$^*$~\cite{hwang2023self}} & {46.21} & {53.21} & {81.71} & {81.82} & {0.7357} & {0.7822} & {53.55} & {55.29} & {85.25} & {84.91} & {0.5389} & {0.7560} \\
    DMD~\cite{li2023decoupled} & 46.06 & - & 83.23 & 83.29 & 0.7520 & - & 52.78 & - & 84.62 & 84.62 & 0.5430 & - \\
    EUAR$^*$~\cite{gao2024enhanced} & 46.06 & 53.21 & 83.54 & 83.59 & \underline{0.7167} & \underline{0.7929} & 52.48 & 54.39 & 83.52 & 83.66 & 0.5363 & 0.7590 \\
    DLF~\cite{wang2025dlf} & 47.08 & 52.33 & \underline{85.06} & \underline{85.04} & 0.7310 & 0.7810 & 53.90 & 55.70 & \underline{85.42} & \underline{85.27} & 0.5360 & 0.7640 \\
    EMOE$^*$~\cite{fang2025emoe} & 46.36 & 54.23 & 83.99 & 83.98 & 0.7281 & 0.7872 & 52.93 & 54.71 & 85.00 & 84.97 & 0.5371 & 0.7641 \\
    CACR$^*$~\cite{wang2026conflict} & 42.86 & 52.19 & 84.60 & 84.55 & 0.7766 & 0.7731 & 53.60 & 55.14 & 84.59 & 84.64 & \underline{0.5356} & 0.7673 \\
    MRUF$^*$~\cite{ma2026mruf} & 46.06 & 53.06 & 84.91 & 84.90 & 0.7170 & 0.7896 & 51.79 & 53.04 & 85.33 & 85.19 & 0.5460 & 0.7592 \\
    \midrule
    \textbf{\method{} (ours)} & \textbf{47.81} & \textbf{55.10} & \textbf{86.59} & \textbf{86.52} & \textbf{0.7067} & \textbf{0.8074} & \textbf{54.73} & \textbf{56.58} & \textbf{86.10} & \textbf{86.05} & \textbf{0.5248} & \textbf{0.7761} \\
    \midrule
    & \multicolumn{6}{c}{CH-SIMS} & \multicolumn{6}{c}{CH-SIMS v2} \\
    \cmidrule(lr){2-7}\cmidrule(lr){8-13}
    Method & Acc-5$\uparrow$ & Acc-3$\uparrow$ & Acc-2$\uparrow$ & F1$\uparrow$ & MAE$\downarrow$ & Corr$\uparrow$
    & Acc-5$\uparrow$ & Acc-3$\uparrow$ & Acc-2$\uparrow$ & F1$\uparrow$ & MAE$\downarrow$ & Corr$\uparrow$ \\
    \midrule
    MulT~\cite{tsai2019multimodal} & 37.94 & 64.77 & \underline{78.56} & \underline{79.66} & 0.4532 & 0.5641 & \underline{54.81} & \underline{73.19} & \underline{80.68} & \underline{80.73} & \underline{0.2905} & \underline{0.7378} \\
    Self-MM$^*$~\cite{yu2021learning} & 42.23 & 64.33 & 76.37 & 76.74 & 0.4378 & 0.5654 & 52.80 & 71.86 & 78.43 & 78.52 & 0.3226 & 0.6602 \\
    {SUGRM$^*$~\cite{hwang2023self}} & {41.58} & {63.89} & {78.12} & {77.88} & {0.4445} & {0.5415} & {45.84} & {65.18} & {75.73} & {75.77} & {0.3424} & {0.6156} \\
    DMD$^*$~\cite{li2023decoupled} & 39.17 & 62.58 & 75.05 & 75.24 & 0.4416 & 0.5434 & 49.42 & 71.47 & 77.18 & 77.28 & 0.3279 & 0.6719 \\
    EUAR$^*$~\cite{gao2024enhanced} & \underline{43.54} & 65.43 & 77.24 & 77.29 & 0.4269 & 0.5757 & 52.03 & 71.57 & 76.69 & 76.72 & 0.3369 & 0.6378 \\
    DLF$^*$~\cite{wang2025dlf} & 37.42 & 64.33 & 77.46 & 77.65 & 0.4473 & 0.5553 & 49.61 & 70.89 & 77.37 & 77.37 & 0.3360 & 0.6605 \\
    EMOE$^*$~\cite{fang2025emoe} & 41.14 & 64.55 & 78.34 & 78.63 & \underline{0.4222} & \underline{0.5803} & 52.03 & 70.21 & 77.18 & 77.26 & 0.3254 & 0.6725 \\
    CACR$^*$~\cite{wang2026conflict} & 21.23 & 54.27 & 69.37 & 56.82 & 0.5879 & 0.1364 & 48.07 & 70.12 & 76.11 & 76.20 & 0.3636 & 0.6398 \\
    MRUF$^*$~\cite{ma2026mruf} & 43.11 & \underline{65.86} & 78.12 & 78.39 & 0.4320 & 0.5774 & 44.49 & 69.15 & 76.40 & 76.44 & 0.3740 & 0.6148 \\
    \midrule
    \textbf{\method{} (ours)} & \textbf{50.11} & \textbf{74.18} & \textbf{85.78} & \textbf{85.35} & \textbf{0.3632} & \textbf{0.7052} & \textbf{58.90} & \textbf{77.27} & \textbf{85.20} & \textbf{85.17} & \textbf{0.2627} & \textbf{0.7878} \\
    \bottomrule
  \end{tabular}%
  }
\end{table*}

\subsection{Datasets}

\textbf{CMU-MOSI}~\cite{zadeh2016multimodal} and \textbf{CMU-MOSEI}~\cite{zadeh2018multimodal} contain 2,199 and 22,856 English video segments with sentiment scores from $-3$ to $3$. \textbf{CH-SIMS}~\cite{yu2020ch} and \textbf{CH-SIMS v2}~\cite{liu2022make} contain 2,281 and 4,403 Chinese video segments scored from $-1$ to $1$. The two Chinese datasets also provide independent unimodal annotations for text, audio, and vision, which we use to define semantic subsets for evaluating UMC under semantically consistent and inconsistent conditions.

\subsection{Experimental Settings}

\subsubsection{Evaluation Protocols}

Following previous work~\cite{tsai2019multimodal}, we report Acc-7, Acc-5, neutral-excluded Acc-2, weighted F1, MAE, and Corr on CMU-MOSI and CMU-MOSEI. On CH-SIMS and CH-SIMS v2, following previous work~\cite{yu2021learning}, we report Acc-5, Acc-3, non-positive versus positive Acc-2, weighted F1, MAE, and Corr. Higher is better except for MAE. For CH-SIMS and CH-SIMS v2, we partition each test set into $D_{\mathrm{msc}}$, whose unimodal and multimodal annotation polarities all agree, and its modality-semantically inconsistent complement $D_{\mathrm{msi}}$.

\subsubsection{Implementation Details}

Our model is trained on one RTX 4090D GPU with Adam~\cite{kingma2014adam}, a nominal batch size of 64, and seed 1111 for at most 100 epochs with patience 10. The task-specific and BERT learning rates are $10^{-3}$ and $5\times10^{-5}$. BERT is frozen in CPLR and fine-tuned in RPC. For feature extraction, text uses BERT-Base~\cite{devlin2019bert}. Audio uses LibROSA~\cite{mcfee2015librosa}, OpenSMILE~\cite{eyben2010opensmile}, or COVAREP~\cite{degottex2014covarep}, while vision uses OpenFace~\cite{baltruvsaitis2016openface} or Facet features~\cite{zadeh2016multimodal,zadeh2018multimodal}. All branches output $d_h=32$ features. CPLR uses five anchors on the CH datasets and seven on the CMU datasets with $\zeta=0.999$. RPC uses $T_r=1$. We employ Optuna~\cite{akiba2019optuna} for hyperparameter optimization.

\subsection{Baselines}

We evaluate UMC by comparing it with representative models in MSA. {We group these baselines into four categories according to their main mechanisms.}

\begin{itemize}
  \item {\textbf{Multi\-modal representation learning and fusion:}\newline MulT~\cite{tsai2019multimodal}, DMD~\cite{li2023decoupled}, and DLF~\cite{wang2025dlf}.}
  \item {\textbf{Pseudo-label generation:}\newline Self-MM~\cite{yu2021learning} and SUGRM~\cite{hwang2023self}.}
  \item {\textbf{Semantic inconsistency modeling:} CACR~\cite{wang2026conflict}.}
  \item {\textbf{Expert coordination and uncertainty-aware fusion:} EMOE~\cite{fang2025emoe}, EUAR~\cite{gao2024enhanced}, and MRUF~\cite{ma2026mruf}.}
\end{itemize}

Notably, methods without public code or required data cannot be compared directly.

\begin{table*}[t]
  \centering
  \caption{The ablation experiments on the CMU-MOSI, CMU-MOSEI, CH-SIMS and CH-SIMS v2 datasets. Bold and underlined values are best and second within each dataset.}
  \label{tab:module-ablation-main}

  \resizebox{\textwidth}{!}{%
  \begin{tabular}{lrrrrrrrrrrrr}
    \toprule
    & \multicolumn{6}{c}{CMU-MOSI} & \multicolumn{6}{c}{CMU-MOSEI} \\
    \cmidrule(lr){2-7}\cmidrule(lr){8-13}
    Variant & Acc-7$\uparrow$ & Acc-5$\uparrow$ & Acc-2$\uparrow$ & F1$\uparrow$ & MAE$\downarrow$ & Corr$\uparrow$ & Acc-7$\uparrow$ & Acc-5$\uparrow$ & Acc-2$\uparrow$ & F1$\uparrow$ & MAE$\downarrow$ & Corr$\uparrow$ \\
    \midrule
    \method{} & \textbf{47.81} & \textbf{55.10} & \textbf{86.59} & \textbf{86.52} & \textbf{0.7067} & \textbf{0.8074} & \textbf{54.73} & \textbf{56.58} & \textbf{86.10} & \textbf{86.05} & \textbf{0.5248} & \textbf{0.7761} \\
    w/o \cplr{} & \underline{40.23} & \underline{51.60} & \underline{84.91} & \underline{84.85} & 0.7850 & \underline{0.7742} & \underline{52.41} & \underline{55.46} & \underline{85.20} & \underline{85.16} & \underline{0.5403} & \underline{0.7690} \\
    w/o \rpc{} & 22.89 & 23.62 & 82.62 & 82.70 & \underline{0.7511} & 0.7177 & 49.24 & 47.93 & 81.37 & \underline{80.91} & 0.6428 & 0.6973 \\
    w/o \cplr{} and \rpc{} & 24.93 & 23.91 & 80.49 & 80.52 & 0.7719 & 0.7006 & 49.26 & 47.39 & 81.73 & \underline{80.91} & 0.6265 & 0.7228 \\
    \midrule
    & \multicolumn{6}{c}{CH-SIMS} & \multicolumn{6}{c}{CH-SIMS v2} \\
    \cmidrule(lr){2-7}\cmidrule(lr){8-13}
    Variant & Acc-5$\uparrow$ & Acc-3$\uparrow$ & Acc-2$\uparrow$ & F1$\uparrow$ & MAE$\downarrow$ & Corr$\uparrow$ & Acc-5$\uparrow$ & Acc-3$\uparrow$ & Acc-2$\uparrow$ & F1$\uparrow$ & MAE$\downarrow$ & Corr$\uparrow$ \\
    \midrule
    \method{} & \textbf{50.11} & \textbf{74.18} & \textbf{85.78} & \textbf{85.35} & \textbf{0.3632} & \textbf{0.7052} & \textbf{58.90} & \textbf{77.27} & \textbf{85.20} & \textbf{85.17} & \textbf{0.2627} & \textbf{0.7878} \\
    w/o \cplr{} & \underline{43.98} & 66.30 & 80.31 & 80.61 & \underline{0.3688} & 0.6081 & \underline{55.80} & \underline{74.66} & \underline{83.56} & \underline{83.50} & \underline{0.2688} & \underline{0.7711} \\
    w/o \rpc{} & 41.14 & 67.83 & \underline{82.06} & \underline{81.22} & 0.3827 & 0.6551 & 51.84 & 69.83 & 80.66 & 80.74 & 0.3043 & 0.7252 \\
    w/o \cplr{} and \rpc{} & 41.79 & \underline{68.27} & 79.43 & 79.55 & 0.3952 & \underline{0.6600} & 51.35 & 67.41 & 80.85 & 80.94 & 0.2936 & 0.7287 \\
    \bottomrule
  \end{tabular}%
  }
\end{table*}

\begin{table}[t]
  \centering
  \caption{Pseudo-label recovery on $D_{\mathrm{msi}}$. M denotes modality. $A_k$ denotes Acc-$k$. Raw denotes the multimodal label. T, A, and V denote text, audio, and vision. Bold values rank first.}
  \label{tab:cplr-recovery-main}

  \resizebox{\columnwidth}{!}{%
  \begin{tabular}{p{1.3cm}lcrrrrrr}
    \toprule
    Dataset & M & Target & A5$\uparrow$ & A3$\uparrow$ & A2$\uparrow$ & F1$\uparrow$ & MAE$\downarrow$ & Corr$\uparrow$ \\
    \midrule
    \multirow{6}{=}{CH-SIMS} & T & Raw & 27.98 & 39.45 & \textbf{65.60} & \textbf{66.93} & 0.4697 & 0.2071 \\
     & & CPLR & \textbf{30.28} & \textbf{39.91} & 62.84 & 64.97 & \textbf{0.3915} & \textbf{0.3156} \\
     & A & Raw & 38.07 & \textbf{55.05} & \textbf{72.94} & \textbf{72.69} & 0.3413 & 0.6217 \\
     & & CPLR & \textbf{39.45} & \textbf{55.05} & 71.10 & 71.29 & \textbf{0.3183} & \textbf{0.6299} \\
     & V & Raw & \textbf{38.99} & 55.50 & 72.02 & 71.52 & \textbf{0.3872} & 0.5861 \\
     & & CPLR & 34.40 & \textbf{56.88} & \textbf{73.39} & \textbf{73.10} & 0.3879 & \textbf{0.6319} \\
    \midrule
    \multirow{6}{=}{CH-SIMS v2} & T & Raw & 40.40 & 47.17 & 63.62 & 64.76 & 0.3484 & 0.4110 \\
     & & CPLR & \textbf{41.50} & \textbf{47.90} & \textbf{66.36} & \textbf{67.40} & \textbf{0.2990} & \textbf{0.5171} \\
     & A & Raw & 57.40 & \textbf{63.25} & \textbf{78.06} & \textbf{78.28} & 0.2388 & \textbf{0.6928} \\
     & & CPLR & \textbf{57.95} & \textbf{63.25} & 75.69 & 75.92 & \textbf{0.2280} & 0.6888 \\
     & V & Raw & \textbf{46.80} & \textbf{52.65} & \textbf{72.76} & 73.37 & 0.2863 & 0.5612 \\
     & & CPLR & 46.44 & \textbf{52.65} & \textbf{72.76} & \textbf{73.39} & \textbf{0.2608} & \textbf{0.6071} \\
    \bottomrule
  \end{tabular}%
  }
\end{table}

\subsection{Performance Comparison}

Table~\ref{tab:main-results} presents the comparison with existing MSA models. UMC ranks first for every reported metric. It exceeds DLF by 1.53\% in Acc-2 on CMU-MOSI and {Self-MM by 0.53\%} in Acc-7 on CMU-MOSEI. It also raises CH-SIMS Acc-3 from {65.86\%} to 74.18\% and CH-SIMS v2 Acc-5 from 54.81\% to 58.90\%, while obtaining the lowest MAE and highest Corr on all four datasets.

These results demonstrate that UMC improves both classification and regression performance across four benchmarks. We believe this is because UMC benefits from the modality-specific supervision of CPLR and the reliability-aware fusion of RPC, which preserve sentiment cues while suppressing uncertain or conflicting predictions.

\subsection{Ablation Study}
\label{sec:controlled-components}

To further investigate the effectiveness of UMC, we conduct three ablations: 1) removing CPLR and using the raw multimodal label, 2) removing RPC and using the arithmetic mean, and 3) removing both components.

As shown in Table~\ref{tab:module-ablation-main}, removing either component degrades all metrics. Without CPLR, CH-SIMS Acc-3 falls from 74.18\% to 66.30\%, showing that a shared multimodal label does not provide the same modality-specific supervision as the recovered targets. Without RPC, CMU-MOSI Acc-7 falls from 47.81\% to 22.89\% and MAE increases on every dataset, showing that the arithmetic mean cannot account for sample-dependent expert quality. Removing both components also underperforms full UMC in every metric.

These findings support the comple\-mentary design of UMC. \newline Specifically, CPLR refines uni\-modal training targets, while RPC determines each expert's contribution during fusion. Appendix~\ref{app:internal-signal-ablation} further ablates the internal signals of both components.

\begin{table}[t]
  \centering
  \caption{Comparison of continuous and discrete CPLR. A$k$ denotes Acc-$k$. Bold values rank first.}
  \label{tab:cplr-continuous-main}

  \resizebox{\columnwidth}{!}{%
  \begin{tabular}{llrrrrrr}
    \toprule
    Dataset & Variant & A7$\uparrow$ & A5$\uparrow$ & A2$\uparrow$ & F1$\uparrow$ & MAE$\downarrow$ & Corr$\uparrow$ \\
    \midrule
    CMU-MOSI & Continuous & \textbf{47.81} & \textbf{55.10} & \textbf{86.59} & \textbf{86.52} & \textbf{0.7067} & \textbf{0.8074} \\
     & Discrete & 43.73 & 51.90 & 83.38 & 83.47 & 0.7533 & 0.7931 \\
    CMU-MOSEI & Continuous & \textbf{54.73} & \textbf{56.58} & \textbf{86.10} & \textbf{86.05} & \textbf{0.5248} & \textbf{0.7761} \\
     & Discrete & 54.02 & 54.45 & 84.65 & 84.67 & 0.5575 & 0.7595 \\
    \midrule
    Dataset & Variant & A5$\uparrow$ & A3$\uparrow$ & A2$\uparrow$ & F1$\uparrow$ & MAE$\downarrow$ & Corr$\uparrow$ \\
    \midrule
    CH-SIMS & Continuous & \textbf{50.11} & \textbf{74.18} & \textbf{85.78} & \textbf{85.35} & \textbf{0.3632} & \textbf{0.7052} \\
     & Discrete & 45.95 & 70.24 & 79.43 & 79.84 & 0.3770 & 0.6272 \\
    CH-SIMS v2 & Continuous & \textbf{58.90} & \textbf{77.27} & \textbf{85.20} & \textbf{85.17} & \textbf{0.2627} & \textbf{0.7878} \\
     & Discrete & 57.45 & 73.69 & 81.24 & 81.34 & 0.2764 & 0.7410 \\
    \bottomrule
  \end{tabular}%
  }
\end{table}

\subsection{CPLR Pseudo-Label Analysis}
\label{sec:cplr-main}

To gain a deeper understanding of CPLR, we compare its recovered pseudo labels with the raw multimodal labels on $D_{\mathrm{msi}}$. Table~\ref{tab:cplr-recovery-main} reports classification and regression metrics between these two types of labels and the ground-truth unimodal annotations provided by CH-SIMS and CH-SIMS v2.

As shown in Table~\ref{tab:cplr-recovery-main}, the recovered CPLR labels outperform the raw multimodal labels in 22 of the 36 metric comparisons and tie them in 4. However, 10 comparisons favor Raw. We speculate that this is because inaccurate predictive means or prototype estimates may move already accurate raw labels away from the unimodal targets, particularly near sentiment-category boundaries.

These results show that CPLR can recover targets closer to the unimodal annotations before fusion in most cases. Appendices~\ref{app:cplr-gate} and~\ref{app:cplr-semantic-boundary} further analyze the confidence-conflict update gate and the effects of CPLR on $D_{\mathrm{msc}}$.

\subsection{Continuous versus Discrete CPLR}
\label{sec:cplr-continuous-main}

To examine whether preserving fine-grained sentiment intensity provides more effective unimodal supervision, we compare the continuous pseudo-labels recovered by CPLR with a discrete variant that maps each pseudo-label to the nearest discrete sentiment level.

As shown in Table~\ref{tab:cplr-continuous-main}, continuous CPLR outperforms discrete CPLR in all 24 comparisons. Specifically, on CH-SIMS, Acc-2 rises from 79.43\% to 85.78\%, MAE falls from 0.3770 to 0.3632, and Corr rises from 0.6272 to 0.7052.

These results show that retaining continuous sentiment values preserves fine-grained sentiment information, further supporting the use of continuous pseudo-labels in CPLR.

\begin{table}[t]
  \centering
  \caption{RPC versus the arithmetic mean on the full test set and highest-disagreement quartile (Q4). A$k$ denotes Acc-$k$. Bold values rank first.}
  \label{tab:rpc-conflict-main}

  \resizebox{\columnwidth}{!}{%
  \begin{tabular}{llrrrrrrrrrrrr}
    \toprule
    \multicolumn{2}{l}{Dataset} & \multicolumn{6}{c}{MOSI} & \multicolumn{6}{c}{MOSEI} \\
    \cmidrule(lr){3-8}\cmidrule(lr){9-14}
    Set & Fuse & A7$\uparrow$ & A5$\uparrow$ & A2$\uparrow$ & F1$\uparrow$ & MAE$\downarrow$ & Corr$\uparrow$ & A7$\uparrow$ & A5$\uparrow$ & A2$\uparrow$ & F1$\uparrow$ & MAE$\downarrow$ & Corr$\uparrow$ \\
    \midrule
    All & Mean & 22.30 & 23.47 & 78.05 & 78.13 & 1.0723 & 0.7526 & 47.61 & 48.47 & 81.10 & 80.81 & 0.6448 & 0.6988 \\
     & RPC & \textbf{47.81} & \textbf{55.10} & \textbf{86.59} & \textbf{86.52} & \textbf{0.7067} & \textbf{0.8074} & \textbf{54.73} & \textbf{56.58} & \textbf{86.10} & \textbf{86.05} & \textbf{0.5248} & \textbf{0.7761} \\
    Q4 & Mean & 10.47 & 10.47 & 80.12 & 81.00 & 1.6538 & 0.7594 & 31.33 & 30.56 & 86.01 & 85.93 & 0.9327 & 0.7831 \\
     & RPC & \textbf{48.84} & \textbf{68.02} & \textbf{92.40} & \textbf{91.91} & \textbf{0.7184} & \textbf{0.8614} & \textbf{51.50} & \textbf{55.36} & \textbf{90.51} & \textbf{90.43} & \textbf{0.5793} & \textbf{0.8600} \\
    \midrule
    \multicolumn{2}{l}{Dataset} & \multicolumn{6}{c}{SIMS} & \multicolumn{6}{c}{SIMS v2} \\
    \cmidrule(lr){3-8}\cmidrule(lr){9-14}
    Set & Fuse & A5$\uparrow$ & A3$\uparrow$ & A2$\uparrow$ & F1$\uparrow$ & MAE$\downarrow$ & Corr$\uparrow$ & A5$\uparrow$ & A3$\uparrow$ & A2$\uparrow$ & F1$\uparrow$ & MAE$\downarrow$ & Corr$\uparrow$ \\
    \midrule
    All & Mean & 37.86 & 66.30 & 80.96 & 78.97 & 0.4116 & 0.6750 & 50.77 & 65.86 & 82.69 & 82.58 & 0.3289 & 0.7578 \\
     & RPC & \textbf{50.11} & \textbf{74.18} & \textbf{85.78} & \textbf{85.35} & \textbf{0.3632} & \textbf{0.7052} & \textbf{58.90} & \textbf{77.27} & \textbf{85.20} & \textbf{85.17} & \textbf{0.2627} & \textbf{0.7878} \\
    Q4 & Mean & 29.57 & 59.13 & 61.74 & 62.47 & 0.5765 & 0.5718 & 34.75 & 44.79 & 73.75 & 73.76 & 0.4569 & 0.6872 \\
     & RPC & \textbf{49.57} & \textbf{80.87} & \textbf{79.13} & \textbf{78.83} & \textbf{0.4382} & \textbf{0.6347} & \textbf{50.58} & \textbf{74.90} & \textbf{81.85} & \textbf{81.64} & \textbf{0.2772} & \textbf{0.7663} \\
    \bottomrule
  \end{tabular}%
  }
\end{table}

\begin{figure}[t]
  \centering
  \includegraphics[width=\columnwidth]{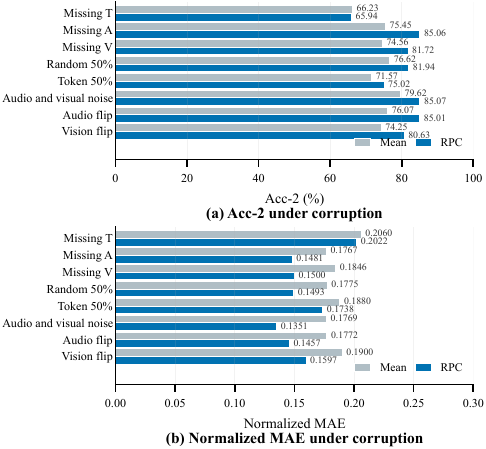}
  \caption{Modality corruption analysis. Normalized MAE is computed by dividing MAE by the dataset's sentiment-label span. Higher Acc-2 and lower normalized MAE are better. Audio and visual noise uses $\sigma_{\mathrm{noise}}=0.5$.}
  \Description{Two horizontal grouped-bar charts compare arithmetic-mean and RPC fusion in absolute Acc-2 and label-span-normalized MAE under eight controlled corruptions.}
  \label{fig:corruption-robustness}
\end{figure}

\subsection{RPC under Prediction Disagreement}
\label{sec:rpc-main}

To further investigate RPC, we compare it with the arithmetic mean using identical expert predictions. Table~\ref{tab:rpc-conflict-main} reports the full test set and its highest prediction-disagreement quartile (Q4).\footnote{Q4 is the top 25\% of samples by label-range-normalized mean absolute difference across the three unimodal prediction pairs.}

As shown in Table~\ref{tab:rpc-conflict-main}, RPC improves all 24 full-test comparisons and all 24 high-disagreement comparisons. Specifically, on CMU-MOSI, full-test Acc-2 rises from 78.05\% to 86.59\%. Especially in Q4, Acc-2 rises from 80.12\% to 92.40\%, while MAE falls from 1.6538 to 0.7184. Overall, every dataset has a larger MAE reduction in Q4 than on its full test set.

Since the expert predictions are fixed within each metric, the gains arise from the fusion rule. Moreover, RPC's larger Q4 gains verify that reliability and precision are especially useful when predictions diverge. Appendices~\ref{app:rpc-fusion-rules} and~\ref{app:rpc-disagreement-strata} further analyze alternative fusion rules and RPC's performance across the full disagreement range.

\begin{figure}[t]
  \centering
  \includegraphics[width=0.95\columnwidth,trim=0 5bp 0 6bp,clip]{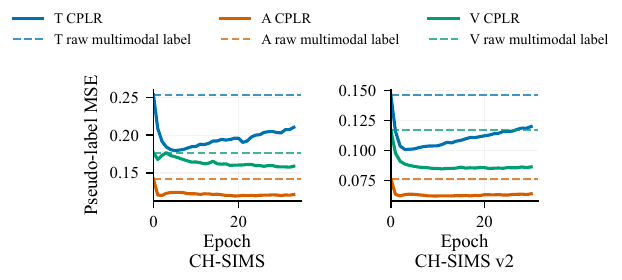}
  \caption{CPLR pseudo-label MSE trajectories. Solid lines show the MSE of CPLR pseudo labels, while dashed lines show the MSE of raw multimodal labels. Lower is better.}
  \Description{Line plots compare text, audio, and visual pseudo-label mean squared error over CPLR epochs with the fixed MSE of raw multimodal labels on CH-SIMS and CH-SIMS v2.}
  \label{fig:cplr-trajectory}
\end{figure}

\subsection{Modality Corruption Analysis}
\label{sec:robustness-main}

To evaluate RPC under corrupted inputs, we apply modality removal, random removal, token deletion, Gaussian noise, and sign reversal. Figure~\ref{fig:corruption-robustness} presents Acc-2 and MAE as representative classification and regression metrics, respectively, and compares RPC with the arithmetic mean using identical experts.

As shown in Figure~\ref{fig:corruption-robustness}, RPC reduces average normalized MAE under all 8 corruptions and improves average Acc-2 under 7, with missing text as the only exception. We speculate that this is because reliability and predictive precision may not sufficiently downweight the text expert when text is missing, allowing its unreliable predictions to affect the final sentiment prediction.

These results show that RPC remains effective under most corruptions. We believe this is because reliability and precision allow RPC to downweight degraded predictions.

\subsection{CPLR Pseudo-Label Update Dynamics}
\label{app:cplr-trajectories}

To examine how CPLR refines modality-specific supervision during optimization,
Figure~\ref{fig:cplr-trajectory} tracks the pseudo-label MSE throughout CPLR
training on CH-SIMS and CH-SIMS v2. For each modality, the MSE values of both
the CPLR pseudo labels and raw multimodal labels are computed against the
ground-truth unimodal annotations.

As shown in Figure~\ref{fig:cplr-trajectory}, the largest adjustments generally
occur during the early epochs, after which the trajectories change gradually.
On both datasets, the text, audio, and visual pseudo-label MSE values end below
their respective raw multimodal label references.

These results show that CPLR reaches stable pseudo-label estimates, with the
consistently lower MSE further verifying the reliability of its generated
pseudo labels.

\section{Conclusion}

We proposed \method{} for multimodal sentiment analysis. First, CPLR recovers continuous modality-specific pseudo-labels through heteroscedastic prediction and ordered sentiment prototypes. RPC then estimates sample-dependent reliability from predictive statistics and cross-modal disagreement and combines it with predictive precision to form the final consensus. Experiments on four public MSA datasets demonstrate the effectiveness of both components and the overall superiority of UMC.

\bibliographystyle{ACM-Reference-Format}
\bibliography{references}

\appendix

\begin{table}[ht]
  \color{black}
  \centering
  \caption{CPLR update strength on $D_{\mathrm{msi}}$. The term $n$ is the number of modality-level instances, and $\bar c$, $\bar\eta$, and $\bar\lambda$ are mean confidence, conflict, and update strength. Both MAEs use held-out unimodal annotations, and $\Delta$MAE is raw minus CPLR. Bold values denote the lower MAE.}
  \label{tab:cplr-gate}

  \resizebox{\columnwidth}{!}{%
  \begin{tabular}{lllrrrrrrr}
    \toprule
    Dataset & Confidence & Conflict & $n$ & $\bar c$ & $\bar\eta$ & $\bar\lambda$ & Raw MAE$\downarrow$ & CPLR MAE$\downarrow$ & $\Delta$MAE \\
    \midrule
    CH-SIMS & Low & Low & 142 & 0.7157 & 0.0694 & 0.0497 & \textbf{0.3352} & 0.3361 & -0.0009 \\
     & Low & High & 185 & 0.7116 & 0.3564 & 0.2531 & 0.4768 & \textbf{0.4231} & +0.0537 \\
     & High & Low & 185 & 0.7498 & 0.0732 & 0.0549 & \textbf{0.2605} & 0.2606 & -0.0001 \\
     & High & High & 142 & 0.7496 & 0.3031 & 0.2270 & 0.5437 & \textbf{0.4583} & +0.0854 \\
    \midrule
    CH-SIMS v2 & Low & Low & 336 & 0.7160 & 0.0792 & 0.0567 & 0.2131 & \textbf{0.2125} & +0.0006 \\
     & Low & High & 485 & 0.7105 & 0.2845 & 0.2018 & 0.3287 & \textbf{0.2983} & +0.0304 \\
     & High & Low & 485 & 0.7603 & 0.0760 & 0.0578 & 0.1996 & \textbf{0.1991} & +0.0005 \\
     & High & High & 335 & 0.7545 & 0.2753 & 0.2078 & 0.4478 & \textbf{0.3531} & +0.0947 \\
    \bottomrule
  \end{tabular}%
  }
\end{table}

\section{CPLR Confidence-Conflict Gate}
\label{app:cplr-gate}

To further examine the CPLR gate, we partition $D_{\mathrm{msi}}$ by
dataset-median confidence and conflict.

As shown in Table~\ref{tab:cplr-gate}, high-con\-flict groups receive update
\newline strengths from $0.2018$ to $0.2531$, compared with $0.0497$ to $0.0567$ in the
low-confidence and low-conflict groups. Their largest MAE reductions reach
$0.0854$ and $0.0947$, while low-conflict changes remain within $0.0009$.

These results show that CPLR can concentrate effective updates where confidence
and label conflict jointly support pseudo-label revision.

\begin{figure}[t]
  \centering
  \includegraphics[width=\columnwidth]{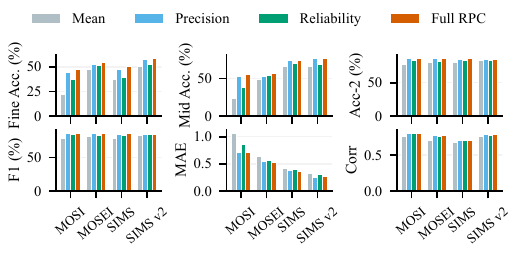}
  \caption{Fixed-expert fusion performance. Fine and mid accuracy denote Acc-7 and Acc-5 on CMU datasets and Acc-5 and Acc-3 on CH datasets. Higher is better except for MAE.}
  \Description{Six grouped bar charts compare arithmetic mean, precision-only fusion, reliability-only fusion, and full RPC with fixed experts on four datasets.}
  \label{fig:rpc-same-experts}
\end{figure}

\section{Fixed-Expert Fusion Rule Comparison}
\label{app:rpc-fusion-rules}

To further isolate RPC from expert training, Figure~\ref{fig:rpc-same-experts}
applies four fusion rules to identical unimodal predictions.

As shown in Figure~\ref{fig:rpc-same-experts}, full RPC improves all 24 metrics
over the arithmetic mean and reliability-only fusion. It also improves 21 of
the 24 metrics over precision-only fusion, with CMU-MOSI Acc-2 and F1 and
CH-SIMS v2 MAE as the 3 exceptions.

These results show that combining reliability and precision can improve
fixed-expert fusion in most settings.

\begin{figure}[t]
  \centering
  \includegraphics[width=0.78\columnwidth]{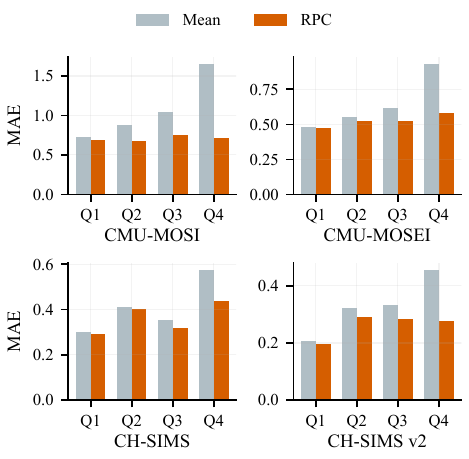}
  \caption{Fixed-expert arithmetic-mean and RPC MAE across prediction-disagreement quartiles. Q1 and Q4 denote the lowest and highest disagreement. Lower is better.}
  \Description{Grouped plots compare RPC and arithmetic-fusion mean absolute error from the lowest to the highest cross-modal disagreement quartile on four datasets.}
  \label{fig:conflict-strata}
\end{figure}

\begin{table}[t]
  \centering
  \caption{RPC signal analysis. Top-1 measures whether the largest reliability $r$, normalized precision $\bar\tau$, or final weight $\omega$ selects the lowest-error expert. The term $\rho_k$ is the reliability correlation with negative modality-$k$ error. Bold and underlined Top-1 values are best and second.}
  \label{tab:router-fidelity-compact}

  \resizebox{\columnwidth}{!}{%
  \begin{tabular}{lrrrrrr}
    \toprule
    & \multicolumn{3}{c}{Top-1 (\%)} & \multicolumn{3}{c}{Reliability correlation} \\
    \cmidrule(lr){2-4}\cmidrule(lr){5-7}
    Dataset & $r$ & $\bar\tau$ & $\omega$ & $\rho_t$ & $\rho_a$ & $\rho_v$ \\
    \midrule
    CMU-MOSI & \textbf{69.53} & \textbf{69.53} & \textbf{69.53} & 0.0633 & 0.5253 & 0.4779 \\
    CMU-MOSEI & \textbf{53.34} & 53.14 & \underline{53.32} & -0.0855 & 0.3979 & 0.3978 \\
    CH-SIMS & \textbf{56.46} & 54.27 & \underline{56.02} & 0.2688 & 0.3534 & 0.1122 \\
    CH-SIMS v2 & \textbf{56.87} & \textbf{56.87} & \underline{55.51} & 0.3488 & 0.2244 & 0.2732 \\
    \bottomrule
  \end{tabular}%
  }
\end{table}

\section{RPC across Prediction-Disagreement Quartiles}
\label{app:rpc-disagreement-strata}

To further determine where RPC helps, Figure~\ref{fig:conflict-strata} compares
fixed-expert RPC with the arithmetic mean. As in the Q4 analysis in
Section~\ref{sec:rpc-main}, we rank samples by the label-range-normalized mean
absolute difference across the three unimodal prediction pairs and divide them
into four quartiles. Q1 contains the bottom 25\% of samples, Q2 contains samples
from the 25th to the 50th percentile, Q3 contains samples from the 50th to the
75th percentile, and Q4 contains the top 25\%.

As shown in Figure~\ref{fig:conflict-strata}, RPC lowers MAE in every
comparison, with each dataset showing its largest reduction in Q4.

These results show that RPC remains useful at low disagreement and contributes
most when unimodal predictions diverge strongly.

\begin{table*}[t]
  \centering
  \caption{Internal-signal ablations. Higher is better except for MAE. Bold and underlined values are best and second. Proto., unc., conf., disag., and sup. abbreviate prototype, uncertainty, conflict, disagreement, and supervision.}
  \label{tab:internal-signal-summary}

  \resizebox{\textwidth}{!}{%
  \begin{tabular}{llrrrrrrrrrrrrrrrrrrrrrrrr}
    \toprule
    & Dataset & \multicolumn{6}{c}{CMU-MOSI} & \multicolumn{6}{c}{CMU-MOSEI} & \multicolumn{6}{c}{CH-SIMS} & \multicolumn{6}{c}{CH-SIMS v2} \\
    \cmidrule(lr){3-8}\cmidrule(lr){9-14}\cmidrule(lr){15-20}\cmidrule(lr){21-26}
    & Metric & Acc-7 & Acc-5 & Acc-2 & F1 & MAE & Corr & Acc-7 & Acc-5 & Acc-2 & F1 & MAE & Corr & Acc-5 & Acc-3 & Acc-2 & F1 & MAE & Corr & Acc-5 & Acc-3 & Acc-2 & F1 & MAE & Corr \\
    \midrule
    Full & \method{} & 47.81 & \textbf{55.10} & \textbf{86.59} & \textbf{86.52} & \textbf{0.7067} & \textbf{0.8074} & \textbf{54.73} & \textbf{56.58} & \textbf{86.10} & \textbf{86.05} & \textbf{0.5248} & \textbf{0.7761} & \textbf{50.11} & \textbf{74.18} & \textbf{85.78} & \textbf{85.35} & \textbf{0.3632} & \textbf{0.7052} & \textbf{58.90} & \textbf{77.27} & \textbf{85.20} & \textbf{85.17} & \textbf{0.2627} & \textbf{0.7878} \\
    \midrule
    \multirow{4}{*}{\shortstack{\cplr{}\\variants}} & Head only & 46.21 & 48.25 & 82.93 & 82.94 & 0.7456 & 0.7592 & 53.68 & 55.53 & 85.36 & 85.24 & 0.5407 & 0.7657 & 40.92 & 66.96 & 80.09 & 80.57 & 0.3867 & 0.6345 & 55.80 & 73.40 & 78.14 & 78.27 & 0.2939 & 0.7456 \\
     & Proto. only & 39.65 & 47.81 & 79.88 & 80.00 & 0.7522 & 0.7637 & 53.40 & 54.99 & 84.81 & 84.72 & 0.5359 & 0.7684 & \underline{45.08} & 68.71 & 79.65 & 80.01 & 0.3762 & 0.6176 & 54.64 & 74.27 & 81.53 & 81.64 & 0.2772 & 0.7616 \\
     & No unc. & 41.40 & 45.63 & 83.38 & 83.36 & 0.7329 & 0.7818 & 53.94 & 54.43 & 85.33 & 85.31 & 0.5335 & \underline{0.7723} & 42.89 & 63.68 & 78.12 & 78.20 & 0.3738 & 0.6300 & 56.67 & 75.73 & 81.24 & 81.34 & 0.2848 & 0.7569 \\
     & No conf. & 23.76 & 23.76 & 85.21 & 85.16 & 0.7338 & 0.7541 & 46.02 & 50.91 & 84.20 & 83.81 & 0.6232 & 0.7211 & 40.48 & 61.93 & 77.46 & 77.69 & 0.3912 & 0.6606 & 50.97 & 66.15 & 80.95 & 81.00 & 0.2737 & 0.7475 \\
    \midrule
    \multirow{5}{*}{\shortstack{\rpc{}\\variants}} & Precision only & 43.73 & 43.59 & 82.62 & 82.72 & 0.7395 & 0.7916 & 52.63 & 53.44 & 85.11 & 85.10 & 0.5464 & 0.7607 & 42.45 & \underline{71.77} & 80.96 & 80.86 & 0.3809 & 0.6809 & \underline{58.61} & 74.56 & 83.75 & 83.82 & 0.2759 & 0.7480 \\
     & Reliability only & 42.13 & 44.90 & 83.84 & 83.76 & 0.7493 & 0.7892 & 52.93 & 56.04 & 85.00 & 84.88 & 0.5417 & 0.7507 & 42.01 & 67.61 & \underline{85.12} & \underline{84.81} & \underline{0.3668} & 0.6691 & 56.00 & 73.02 & 78.72 & 78.84 & 0.2973 & 0.7568 \\
     & No unc. & 45.34 & \underline{50.58} & \underline{85.67} & 85.56 & \underline{0.7175} & 0.7906 & \underline{54.30} & 56.47 & 85.66 & 85.55 & 0.5307 & 0.7708 & 39.82 & 68.71 & 80.09 & 80.25 & 0.3895 & \underline{0.6811} & 56.58 & 75.24 & 83.46 & 83.54 & 0.2738 & 0.7824 \\
     & No disag. & \textbf{48.69} & 49.56 & 84.76 & 84.80 & 0.7593 & \underline{0.7988} & \textbf{54.73} & \underline{56.49} & \underline{85.69} & \underline{85.61} & \underline{0.5287} & 0.7717 & 44.86 & 68.71 & 81.62 & 81.76 & 0.3732 & 0.6559 & 56.38 & 76.21 & \underline{84.72} & \underline{84.71} & 0.2756 & \underline{0.7877} \\
     & No sup. & \underline{48.54} & 46.79 & \underline{85.67} & \underline{85.65} & 0.7178 & 0.7846 & 53.72 & 55.33 & 85.20 & 85.03 & 0.5359 & 0.7650 & 41.14 & 68.49 & 83.37 & 83.27 & 0.3829 & 0.6674 & 57.64 & \underline{76.40} & 84.53 & 84.51 & \underline{0.2723} & 0.7818 \\
    \bottomrule
  \end{tabular}%
  }
\end{table*}

\begin{table*}[t]
  \centering
  \caption{Inference efficiency of UMC. Mean uses arithmetic-mean fusion with the same experts, while RPC ms and Sample/s measure complete UMC inference. Params, FLOPs, and $\Delta t$ report the incremental RPC cost. Latency is the median of 100 batches on one RTX 4090D. Bold values denote the lower latency.}
  \label{tab:efficiency}

  \resizebox{0.60\textwidth}{!}{%
  \begin{tabular}{lrrrrrr}
    \toprule
    Dataset & Params & FLOPs & Mean ms & RPC ms & $\Delta t$ & Sample/s \\
    \midrule
    CMU-MOSI    & 2,305 & 13,080 & \textbf{29.2024} & 29.8395 & 2.18\% & 2144.8 \\
    CMU-MOSEI   & 2,305 & 13,080 & \textbf{28.0514} & 28.5647 & 1.83\% & 2240.5 \\
    CH-SIMS     & 2,305 & 13,080 & \textbf{32.2526} & 32.8795 & 1.94\% & 1946.5 \\
    CH-SIMS v2  & 2,305 & 13,080 & \textbf{29.7458} & 30.4773 & 2.46\% & 2099.9 \\
    \bottomrule
  \end{tabular}%
  }
\end{table*}

\section{Reliability, Precision, and Expert Quality}
\label{app:rpc-router-fidelity}

To further examine whether RPC signals reflect expert quality, we test whether
reliability $r$, normalized precision $\bar\tau$, and final fusion weight
$\omega$ identify the most accurate unimodal expert for each test sample. We
first compute the error $e_{k,i}$ of expert $k$ and the lowest-error expert
$b_i$:
\begin{equation}
  e_{k,i}=\left|\mu^{\mathrm{pred}}_{k,i}-y_i^m\right|,
  \qquad
  b_i=\arg\min_{k\in\mathcal{M}}e_{k,i},
\label{eq:expert-quality-reference}
\end{equation}
where $y_i^m$ is the multimodal label and $b_i$ is used as the reference
for expert selection.

We next calculate normalized precision $\bar\tau_{k,i}$ and the final fusion
weight $\omega_{k,i}$ corresponding to Eq.~\eqref{eq:consensus-prediction}:
\begin{equation}
  \bar\tau_{k,i}
  =\frac{\tau_{k,i}}{\sum_{q\in\mathcal{M}}\tau_{q,i}},
  \qquad
  \omega_{k,i}
  =\frac{r_{k,i}\tau_{k,i}}
  {\sum_{q\in\mathcal{M}}r_{q,i}\tau_{q,i}},
\label{eq:diagnostic-fusion-signals}
\end{equation}
where both normalizers sum over text, audio, and vision.

Using these quantities, we calculate the Top-1 score $A_{\mathrm{top1}}(s)$ for
each signal $s\in\{r,\bar\tau,\omega\}$ and the modality-wise Spearman
correlation $\rho_k$ for reliability:
\begin{equation}
  \begin{aligned}
  A_{\mathrm{top1}}(s)
    &=\frac{100}{N}\sum_{i=1}^{N}
      \mathbf{1}\!\left[\arg\max_{k\in\mathcal{M}}s_{k,i}=b_i\right],\\
  \rho_k
    &=\operatorname{Spearman}\!\left(
      (r_{k,i})_{i=1}^{N},(-e_{k,i})_{i=1}^{N}\right),
  \end{aligned}
\label{eq:diagnostic-quality-alignment}
\end{equation}
where $N$ is the number of test samples and $\mathbf{1}[\cdot]$ is the indicator
function. A positive $\rho_k$ indicates that higher reliability is associated
with lower error.

As shown in Table~\ref{tab:router-fidelity-compact}, all three signal types
select the lowest-error expert at rates from 53.14\% to 69.53\%, exceeding the
33.33\% expected from random selection. Reliability is best or tied on every
dataset, and 11 of its 12 modality-wise correlations are positive.

These results show that reliability and precision both carry expert-quality
information, supporting their joint use in RPC to adaptively weight unimodal
experts during fusion.

\begin{table}[t]
  \centering
  \caption{Pseudo-label recovery on $D_{\mathrm{msc}}$. Raw denotes the multimodal label. T, A, and V denote text, audio, and vision. Bold values rank first.}
  \label{tab:cplr-subset-boundary}

  \resizebox{\columnwidth}{!}{%
  \begin{tabular}{p{1.3cm}lcrrrrrr}
    \toprule
    Dataset & M & Target & A5$\uparrow$ & A3$\uparrow$ & A2$\uparrow$ & F1$\uparrow$ & MAE$\downarrow$ & Corr$\uparrow$ \\
    \midrule
    \multirow{6}{=}{CH-SIMS} & T & Raw & \textbf{71.55} & \textbf{100.00} & \textbf{100.00} & \textbf{100.00} & \textbf{0.1908} & 0.9267 \\
     & & CPLR & 70.71 & 99.58 & 99.16 & 99.17 & 0.2000 & \textbf{0.9328} \\
     & A & Raw & \textbf{75.73} & \textbf{100.00} & \textbf{100.00} & \textbf{100.00} & \textbf{0.1506} & 0.9411 \\
     & & CPLR & 62.76 & \textbf{100.00} & 97.91 & 97.93 & 0.2335 & \textbf{0.9429} \\
     & V & Raw & \textbf{80.33} & \textbf{100.00} & \textbf{100.00} & \textbf{100.00} & \textbf{0.1238} & \textbf{0.9557} \\
     & & CPLR & 73.22 & 99.16 & \textbf{100.00} & \textbf{100.00} & 0.2226 & 0.9520 \\
    \midrule
    \multirow{6}{=}{CH-SIMS v2} & T & Raw & 65.09 & \textbf{100.00} & \textbf{100.00} & \textbf{100.00} & 0.1791 & 0.9313 \\
     & & CPLR & \textbf{71.25} & \textbf{100.00} & 97.33 & 97.35 & \textbf{0.1492} & \textbf{0.9420} \\
     & A & Raw & \textbf{78.85} & \textbf{100.00} & \textbf{100.00} & \textbf{100.00} & \textbf{0.1195} & \textbf{0.9612} \\
     & & CPLR & 75.36 & 99.79 & 96.51 & 96.54 & 0.1516 & 0.9575 \\
     & V & Raw & 75.15 & \textbf{100.00} & \textbf{100.00} & \textbf{100.00} & \textbf{0.1409} & \textbf{0.9531} \\
     & & CPLR & \textbf{78.44} & 99.59 & 99.59 & 99.59 & 0.1413 & 0.9495 \\
    \bottomrule
  \end{tabular}%
  }
\end{table}

\section{Internal Signal Ablation}
\label{app:internal-signal-ablation}

To examine CPLR and RPC components, Table~\ref{tab:internal-signal-summary}
compares controlled UMC variants. CPLR retains only the regression head (Head
only) or prototype branch (Proto. only), or omits uncertainty weighting (No
unc.) or conflict gating (No conf.). RPC retains only precision (Precision
only) or reliability (Reliability only), or omits uncertainty input (No unc.),
disagreement (No disag.), or router supervision (No sup.).

As shown in Table~\ref{tab:internal-signal-summary}, full UMC wins or ties 23 of
24 ablation comparisons. These results further demonstrate the importance of
the individual components in CPLR and RPC, as removing or isolating any
component degrades performance in most cases.

\begin{table*}[t]
  \centering
  
  \caption{{Component ablations on modality-semantically consistent ($D_{\mathrm{msc}}$) and inconsistent ($D_{\mathrm{msi}}$) test subsets. Higher is better except for MAE. Bold and underlined values are best and second. Proto., unc., conf., disag., and sup. abbreviate prototype, uncertainty, conflict, disagreement, and supervision.}}
  \label{tab:component-semantic-subsets}

  \resizebox{\textwidth}{!}{%
  \begin{tabular}{llrrrrrrrrrrrrrrrrrrrrrrrr}
    \toprule
    & Dataset & \multicolumn{12}{c}{CH-SIMS} & \multicolumn{12}{c}{CH-SIMS v2} \\
    \cmidrule(lr){3-14}\cmidrule(lr){15-26}
    & Subset & \multicolumn{6}{c}{$D_{\mathrm{msc}}$} & \multicolumn{6}{c}{$D_{\mathrm{msi}}$} & \multicolumn{6}{c}{$D_{\mathrm{msc}}$} & \multicolumn{6}{c}{$D_{\mathrm{msi}}$} \\
    \cmidrule(lr){3-8}\cmidrule(lr){9-14}\cmidrule(lr){15-20}\cmidrule(lr){21-26}
    & Metric & Acc-5 & Acc-3 & Acc-2 & F1 & MAE & Corr & Acc-5 & Acc-3 & Acc-2 & F1 & MAE & Corr & Acc-5 & Acc-3 & Acc-2 & F1 & MAE & Corr & Acc-5 & Acc-3 & Acc-2 & F1 & MAE & Corr \\
    \midrule
    Full & \method{} & \textbf{56.07} & \textbf{87.45} & \underline{92.47} & \underline{92.20} & 0.3338 & \textbf{0.8368} & \textbf{43.58} & \textbf{59.63} & \textbf{78.44} & \underline{77.92} & 0.3954 & \underline{0.4165} & \textbf{58.73} & \underline{88.30} & \underline{94.05} & \underline{94.05} & \textbf{0.2161} & 0.8749 & \underline{59.05} & \textbf{67.46} & \textbf{77.33} & \textbf{77.27} & \underline{0.3043} & \textbf{0.6281} \\
    \midrule
    \multirow{3}{*}{w/o} & \cplr{} & \underline{52.30} & 79.08 & 89.96 & 90.01 & 0.3377 & 0.7389 & 34.86 & 52.29 & 69.72 & 70.24 & 0.4028 & 0.3114 & 54.83 & 85.01 & 92.81 & 92.80 & \underline{0.2270} & 0.8665 & 56.67 & 65.45 & 75.32 & 75.23 & 0.3060 & 0.5895 \\
     & \rpc{} & 46.86 & 82.01 & 88.70 & 88.17 & 0.3856 & 0.7790 & 34.86 & 52.29 & \underline{74.77} & 73.65 & \textbf{0.3795} & 0.3595 & 50.92 & 81.72 & 90.76 & 90.82 & 0.2733 & 0.8338 & 52.65 & 59.23 & 71.66 & 71.70 & 0.3320 & 0.5050 \\
     & \cplr{} and \rpc{} & 43.93 & 79.50 & 89.12 & 89.00 & 0.4008 & 0.7684 & 39.45 & 55.96 & 68.81 & 69.18 & \underline{0.3890} & 0.3955 & 51.13 & 79.06 & 89.73 & 89.78 & 0.2642 & 0.8337 & 51.55 & 57.04 & 72.94 & 72.96 & 0.3198 & 0.5156 \\
    \midrule
    \multirow{4}{*}{\shortstack{\cplr{}\\variants}} & Head only & 39.75 & 80.33 & 91.63 & 91.67 & 0.3617 & 0.7619 & \underline{42.20} & 52.29 & 67.43 & 68.15 & 0.4142 & 0.3522 & 55.03 & 84.19 & 86.65 & 86.77 & 0.2743 & 0.8544 & 56.49 & 63.80 & 70.57 & 70.52 & 0.3115 & 0.5370 \\
     & Proto. only & 48.54 & 80.33 & 89.96 & 90.05 & 0.3403 & 0.7552 & 41.28 & 55.96 & 68.35 & 68.91 & 0.4154 & 0.3111 & 51.95 & 85.83 & 90.14 & 90.27 & 0.2392 & 0.8735 & 57.04 & 63.99 & 73.86 & 73.83 & 0.3111 & 0.5517 \\
     & No unc. & 43.93 & 79.92 & 89.54 & 89.33 & \underline{0.3325} & 0.7705 & 41.74 & 45.87 & 65.60 & 66.04 & 0.4192 & 0.3215 & 55.44 & 86.45 & 90.97 & 91.06 & 0.2566 & 0.8570 & 57.77 & \underline{66.18} & 72.58 & 72.59 & 0.3099 & 0.5600 \\
     & No conf. & 43.10 & 77.82 & 89.54 & 89.33 & 0.3861 & 0.7908 & 37.61 & 44.50 & 64.22 & 64.87 & 0.3967 & 0.3414 & 51.54 & 80.49 & 91.99 & 92.04 & 0.2388 & 0.8535 & 50.46 & 53.38 & 71.12 & 71.14 & 0.3047 & 0.5421 \\
    \midrule
    \multirow{5}{*}{\shortstack{\rpc{}\\variants}} & Precision only & 47.70 & \underline{85.77} & 90.79 & 90.70 & 0.3579 & 0.7937 & 36.70 & \underline{56.42} & 70.18 & 70.14 & 0.4061 & 0.3985 & \underline{57.08} & 86.65 & 93.02 & 93.05 & 0.2436 & 0.8657 & \textbf{59.96} & 63.80 & 75.50 & 75.53 & 0.3047 & 0.5367 \\
     & Reliability only & 47.70 & 82.43 & 91.21 & 90.86 & \textbf{0.3281} & 0.8135 & 35.78 & 51.38 & \textbf{78.44} & \textbf{78.27} & 0.4092 & 0.3433 & 56.88 & 85.83 & 89.32 & 89.38 & 0.2796 & 0.8562 & 55.21 & 61.61 & 69.29 & 69.26 & 0.3131 & 0.5645 \\
     & No unc. & 39.75 & 82.85 & 90.38 & 90.30 & 0.3828 & 0.7929 & 39.91 & 53.21 & 68.81 & 69.24 & 0.3969 & \textbf{0.4184} & 54.41 & 87.27 & 92.20 & 92.26 & 0.2432 & 0.8734 & 58.50 & 64.53 & 75.69 & 75.72 & \textbf{0.3010} & 0.6167 \\
     & No disag. & 47.70 & 83.26 & 91.21 & 91.28 & 0.3568 & 0.7809 & 41.74 & 52.75 & 71.10 & 71.29 & 0.3912 & 0.3592 & 54.41 & \textbf{88.50} & 93.84 & 93.85 & 0.2425 & \underline{0.8751} & 58.14 & 65.27 & \underline{76.60} & \underline{76.57} & 0.3051 & \underline{0.6276} \\
     & No sup. & 44.35 & 82.43 & \textbf{92.89} & \textbf{92.75} & 0.3637 & \underline{0.8249} & 37.61 & 53.21 & 72.94 & 72.97 & 0.4040 & 0.3553 & 56.26 & 88.09 & \textbf{94.25} & \textbf{94.26} & 0.2317 & \textbf{0.8754} & 58.87 & 66.00 & 75.87 & 75.83 & 0.3085 & 0.6108 \\
    \bottomrule
  \end{tabular}%
  }
\end{table*}

\section{Inference Efficiency}
\label{app:inference-efficiency}

To further assess UMC's deployment cost, we compare RPC with the arithmetic
mean using identical unimodal experts. Table~\ref{tab:efficiency} reports
end-to-end latency and throughput while isolating the parameters, FLOPs, and
latency added by RPC.

As shown in Table~\ref{tab:efficiency}, the router adds 2,305 parameters and
13,080 FLOPs per sample. Although RPC increases median batch latency by
$1.83\%$ to $2.46\%$, throughput remains between 1,946.5 and 2,240.5 samples
per second. Meanwhile, Table~\ref{tab:module-ablation-main} shows that RPC
improves all 24 reported metrics over the arithmetic mean.

These results show that RPC provides consistent performance gains with limited
additional inference cost, yielding a favorable performance-efficiency
trade-off for UMC.

\section{CPLR across Semantic-Consistency Subsets}
\label{app:cplr-semantic-boundary}

To examine whether CPLR gains depend on semantic inconsistency, we further
compare its recovered pseudo labels with the raw multimodal labels on
$D_{\mathrm{msc}}$, complementing the $D_{\mathrm{msi}}$ analysis in
Table~\ref{tab:cplr-recovery-main}. Table~\ref{tab:cplr-subset-boundary}
reports classification and regression metrics between these two types of
labels and the ground-truth unimodal annotations provided by CH-SIMS and
CH-SIMS v2.

As shown in Table~\ref{tab:cplr-subset-boundary}, the recovered CPLR labels
outperform the raw multimodal labels in 6 of the 36 metric comparisons and tie
them in 4. This includes improvements in 4 of the 12 MAE and Corr comparisons.
Specifically, CH-SIMS v2 text MAE falls from 0.1791 to 0.1492, while Corr
increases from 0.9313 to 0.9420.

Together with Table~\ref{tab:cplr-recovery-main}, these results show that CPLR
recovers targets closer to the unimodal annotations more often on
$D_{\mathrm{msi}}$ than on $D_{\mathrm{msc}}$, supporting its role in refining
semantically inconsistent supervision.

\section{{Component Ablation by Semantic Consistency}}
\label{app:semantic-subset-ablation}

{
To further examine the contributions of CPLR and RPC under different levels of
semantic consistency, we evaluate UMC and the variants in
Tables~\ref{tab:module-ablation-main} and~\ref{tab:internal-signal-summary} on
$D_{\mathrm{msc}}$ and $D_{\mathrm{msi}}$ from CH-SIMS and CH-SIMS v2. These
test subsets distinguish whether the unimodal sentiment annotations agree or
disagree, complementing the full-test-set ablations.
\par}

{
As shown in Table~\ref{tab:component-semantic-subsets}, full UMC ranks first or
second in 21 of the 24 metric
comparisons across the two datasets and two subsets. Specifically, on
$D_{\mathrm{msi}}$, removing CPLR lowers Acc-2 from 78.44\% to 69.72\% on
CH-SIMS and from 77.33\% to 75.32\% on CH-SIMS v2. Removing RPC lowers the
corresponding values to 74.77\% and 71.66\%, respectively.
\par}

{
These results show that the complete design remains competitive on
$D_{\mathrm{msc}}$. In particular, the classification gains on
$D_{\mathrm{msi}}$ further support the joint use of modality-specific target
refinement and adaptive consensus when unimodal sentiments are inconsistent.
\par}

\begin{figure}[t]
  \centering
  \includegraphics[width=\columnwidth]{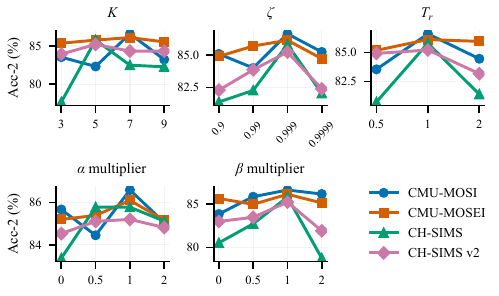}
  \caption{{Acc-2 sensitivity to prototype count $K$, prototype momentum $\zeta$, router target temperature $T_r$, router-loss weight $\alpha$, and unimodal-loss weight $\beta$ on four datasets. The $\alpha$ and $\beta$ axes show multipliers of their selected values.}}
  \Description{Five plots show binary classification accuracy when prototype count, prototype momentum, router target temperature, router-loss weight, or unimodal-loss weight is varied independently on four datasets.}
  \label{fig:sensitivity}
\end{figure}

\section{{Hyperparameter Sensitivity}}
\label{app:hyperparameter-sensitivity}

{
To examine how the main hyperparameters influence UMC, we vary one parameter
at a time for each of the four
datasets, keeping the remaining parameters unchanged. The analysis covers
prototype count $K$, prototype momentum $\zeta$, router target temperature
$T_r$, router-loss weight $\alpha$, and unimodal-loss weight $\beta$.
\par}

{
As shown in Figure~\ref{fig:sensitivity}, the effect of each parameter varies
across datasets. CH-SIMS obtains its highest Acc-2 at $K=5$, whereas CMU-MOSI
and CMU-MOSEI favor $K=7$ among the tested values. Prototype momentum
$\zeta=0.999$ and router target temperature $T_r=1$ give the highest Acc-2 on
all four datasets in their respective sweeps. Changing $\alpha$ produces
smaller variations, while the $\beta$ sweep shows a more pronounced decrease
on CH-SIMS when unimodal regularization is removed or doubled.
\par}

{
These results show that UMC is relatively insensitive to the router-loss
weight $\alpha$. Changes in prototype momentum
$\zeta$ and router target temperature $T_r$ produce moderate,
dataset-dependent variations, whereas prototype count $K$ and unimodal-loss
weight $\beta$ have a stronger effect on CH-SIMS.
\par}

\end{document}